\documentclass{article}

\usepackage[final, nonatbib]{neurips_2023}
\usepackage[square,sort,comma,numbers]{natbib}

\usepackage{fancyhdr}

\fancypagestyle{plain}{%
    \renewcommand{\headrulewidth}{0pt}%
    \fancyhf{}%
    \fancyfoot[R]{\thepage}%
}

\usepackage[utf8]{inputenc} 
\usepackage[T1]{fontenc}    
\usepackage{hyperref}       
\usepackage{url}            
\usepackage{booktabs}       
\usepackage{amsfonts}       
\usepackage{nicefrac}       
\usepackage{microtype}      
\usepackage{xcolor}         
\usepackage[pdftex]{graphicx} 
\usepackage{microtype}
\usepackage{subfigure}
\usepackage{booktabs} 
\usepackage{color,colortbl}
\usepackage[linewidth=1pt]{mdframed}
\usepackage{framed}
\usepackage{adjustbox}

\usepackage{amsmath}
\usepackage{amssymb}
\usepackage{mathtools}
\usepackage{amsthm}

\usepackage[capitalize,noabbrev]{cleveref}

\theoremstyle{plain}

\theoremstyle{definition}

\theoremstyle{remark}

\usepackage[textsize=tiny]{todonotes}

\title{Stabilizing Backpropagation in 16-bit Neural Training with Modified Adam Optimizer}

\author{
  Juyoung Yun\textsuperscript{1,2}\thanks{Corresponding author: \texttt{daniel@open-nn.com}} \\ \\
  \textsuperscript{1}Stony Brook University, Department of Computer Science, United States \\
  \textsuperscript{2}OpenNN Lab, MODULABS, Republic of Korea \\ 
}

\begin{document}

\maketitle

\begin{abstract}
In this research, we address critical concerns related to the numerical instability observed in 16-bit computations of machine learning models. Such instability, particularly when employing popular optimization algorithms like RMSProp and Adam, often leads to unreliable training of deep neural networks. This not only disrupts the learning process but also poses significant challenges in deploying dependable models in real-world applications. Our investigation identifies the epsilon hyperparameter as the primary source of this instability. A nuanced exploration reveals that subtle adjustments to epsilon within 16-bit computations can enhance the reliability of RMSProp and Adam, enabling more stable training of 16-bit neural networks. We propose a novel, dependable approach that leverages updates from the Adam optimizer to bolster the stability of the learning process. Our contributions provide deeper insights into optimization challenges in low-precision computations and offer solutions to ensure the stability of deep neural network training, paving the way for their dependable use in various applications.
\end{abstract}


\section{Introduction}
The meteoric advancement of machine learning and artificial intelligence technologies has enabled the construction of neural networks that effectively emulate the complex computations of the human brain. These deep learning models have found utility in a wide range of applications, such as computer vision, natural language processing, autonomous driving, and more. With the growing complexity and sophistication of these neural network models, the computational requirements, particularly for 32-bit operations, have exponentially increased. This heightened computational demand necessitates the exploration of more efficient alternatives, such as 16-bit operations.

However, the shift to 16-bit operations is riddled with challenges. A common standpoint within the research community argues that 16-bit operations are not ideally suited for neural network computations. This belief is mainly attributable to concerns related to numerical instability during the backpropagation phase, especially when popular optimizers like Adam \cite{kingma2014adam} and RMSProp \cite{tieleman2012lecture} are employed. This instability, more pronounced during the optimizer-mediated backpropagation process rather than forward propagation, can negatively impact the performance of 16-bit operations and compromise the functioning of the neural network model. Current optimizers predominantly operate on 32-bit precision. If these are deployed in a 16-bit environment without appropriate hyperparameter fine-tuning, the neural network models encounter difficulties during learning. This issue is particularly evident in backward propagation, which heavily relies on the optimizer. Confronted with these challenges, the objective of this research is to conduct an exhaustive investigation into the feasibility and implementation of 16-bit operations for neural network training. We propose and evaluate innovative strategies aimed at reducing the numerical instability encountered during the backpropagation phase under 16-bit environments. A significant focus of this paper is also dedicated to exploring the future possibilities of developing 16-bit based optimizers. One of the fundamental aims of this research is to adapt key optimizers such as RMSProp and Adam to prevent numerical instability, thereby facilitating efficient 16-bit computations. These newly enhanced optimizers are designed to not only address the issue of numerical instability but also leverage the computational advantages offered by 16-bit operations, all without compromising the overall performance of the neural network models. Through this research, our intention goes beyond improving the efficiency of neural network training; we also strive to validate the use of 16-bit operations as a dependable and efficient computational methodology in the domain of deep learning. We anticipate that our research will contribute to a shift in the prevalent perceptions about 16-bit operations and will foster further innovation in the field. Ultimately, we hope our findings will pave the way for a new era in deep learning research characterized by efficient, high-performance neural network models.

\section{Related Works}
The matter of numerical precision in deep learning model training has garnered substantial attention in recent years. A milestone study by Gupta et al. \cite{Gupta2015} was among the first to explore the potential of lower numerical precision in deep learning, emphasizing the critical balance between computational efficiency and precision. They suggested that with careful implementation, lower precision models can be as effective as their higher precision counterparts while requiring less computational and memory resources. Building upon these findings, significant advancements have been made in utilizing 16-bit operations for training convolutional neural networks. Courbariaux et al. \cite{Courbariaux2016} pioneered a novel technique to train neural networks using binary weights and activations, significantly reducing memory and computational demands. Despite these advancements, numerical instability during backpropagation remains a persistent challenge. Bengio et al. \cite{Bengio1994} illustrated the difficulties encountered in learning long-term dependencies with gradient descent due to numerical instability. Such findings underscore the need for innovative solutions to mitigate this widespread issue. One critical component to addressing this challenge lies in the development of effective optimizers. Adam, an optimizer introduced by Kingma and Ba \cite{kingma2014adam}, is known to face issues of numerical instability when employed in lower-precision environments. In the context of efficient neural network training, Han et al. \cite{Han2015} proposed a three-stage pipeline that significantly reduces the storage requirements of the network. Their work forms an integral part of the broader discussion on efficient neural network training, further reinforcing the relevance of our research on 16-bit operations. Through our investigation, we aim to contribute to this body of work by presenting a novel approach to addressing numerical instability issues associated with 16-bit precision in the realm of deep learning model training. 

\textbf{Mixed Precision Training}: Micikevicius et al. \cite{Micikevicius2017} stressed the imperative transition from 32-bit to 16-bit operations in deep learning, given the memory and computational constraints associated with training increasingly complex neural networks. Their research demonstrated that mixed precision computations offer a more memory and computation-efficient alternative without compromising the model's performance. Another research by Yun et al. \cite{yun2023defense}, provided a comprehensive theoretical analysis, focusing on the performance of pure 16-bit floating-point neural networks. They introduced the concepts of floating-point error and tolerance to define the conditions under which 16-bit models could approximate their 32-bit counterparts. Their results indicate that pure 16-bit floating-point neural networks can achieve similar or superior performance compared to their mixed-precision and 32-bit counterparts, offering a unique perspective on the benefits of pure 16-bit networks. 

\section{Analysis}
Through theoretical analysis, we analyze which part of the neural network occurs numerical instability which can ruin the training process.

\subsection{Forward Propagation}
\noindent\textbf{Linear Network.} During forward propagation, each column of the input matrix $X$ represents a distinct training example. This matrix layout simplifies the network's computation, as each feature of every training sample can be processed in parallel, exploiting the parallel nature of matrix operations.
\begin{equation*}
X = \begin{bmatrix}
x_{11} & x_{12} & \cdots & x_{1m} \\
x_{21} & x_{22} & \cdots & x_{2m} \\
\vdots & \vdots & \ddots & \vdots \\
x_{n1} & x_{n2} & \cdots & x_{nm}
\end{bmatrix}
\end{equation*}
The weight matrix $W$ represents the strength and direction of connections between neurons. Each column of $W$ corresponds to the set of weights connecting every input neuron to a specific hidden neuron.
\begin{equation*}
W = \begin{bmatrix}
w_{11} & w_{12} & \cdots & w_{1k} \\
w_{21} & w_{22} & \cdots & w_{2k} \\
\vdots & \vdots & \ddots & \vdots \\
w_{n1} & w_{n2} & \cdots & w_{nk}
\end{bmatrix}
\end{equation*}
The bias vector $b$ provides an additional degree of freedom, allowing each neuron in the hidden layer to be activated not just based on weighted input but also based on this inherent property.
\begin{equation*}
b = \begin{bmatrix}
b_{1} \\
b_{2} \\
\vdots \\
b_{k}
\end{bmatrix}
\end{equation*}
The raw outputs or pre-activations (denoted as $Z$) are computed by multiplying the transposed weight matrix with the input matrix and adding the bias. 
\begin{equation*}
Z = W^T X + b
\end{equation*}
After computing $Z$, we need a mechanism to introduce non-linearity into our model. Without this, no matter how deep our model, it would behave just like a single-layer perceptron. This non-linearity is introduced using an activation function $\sigma(\cdot)$, applied element-wise to the matrix $Z$.
\begin{equation*}
A = \sigma(Z)
\end{equation*}
Here, each entry $a_{ij}$ in $A$ is the activated value of the $j$-th neuron in the hidden layer when the model is provided the $i$-th training example. These activations will either be used as input for subsequent layers or be the final output of the network.

In Deep Neural Network (Linear Network.), each operation in the forward pass involves straightforward mathematical operations like addition and multiplication. There is no division or complex operation that could amplify small numerical errors or lead to potential instability. \\

\noindent\textbf{Convolutional Network.} In the realm of deep learning, especially when processing image data, CNNs have gained significant prominence. The foundational blocks of CNNs involve convolving filters over input data and pooling layers \cite{lecun1998gradient}. 

The convolution layer can be detailed by observing the element-wise multiplication and summation \cite{lecun1998gradient}. Specifically, in our 3x3 input matrix \(I\) and 2x2 filter \(F\) example:
\begin{align*}
I &= \begin{bmatrix}
i_{11} & i_{12} & i_{13} \\
i_{21} & i_{22} & i_{23} \\
i_{31} & i_{32} & i_{33}
\end{bmatrix}
&
F &= \begin{bmatrix}
f_{11} & f_{12} \\
f_{21} & f_{22}
\end{bmatrix}
\end{align*}
For the top-left corner of \(I\), the convolution operation using the filter \(F\) is as follows:
\begin{equation*}
o_{11} = \begin{bmatrix}
i_{11} & i_{12} \\
i_{21} & i_{22}
\end{bmatrix} 
\odot 
\begin{bmatrix}
f_{11} & f_{12} \\
f_{21} & f_{22}
\end{bmatrix}
\end{equation*}
Where \(\odot\) represents element-wise multiplication. This implies:
\begin{equation*}
o_{11} = i_{11} \cdot f_{11} + i_{12} \cdot f_{12} + i_{21} \cdot f_{21} + i_{22} \cdot f_{22}
\end{equation*}
The filter \(F\) continues sliding across \(I\), performing similar computations for each position, resulting in the matrix:
\begin{equation*}
O = \begin{bmatrix}
o_{11} & o_{12} \\
o_{21} & o_{22}
\end{bmatrix}
\end{equation*}
Each element in the output matrix \(O\) is computed in the same fashion, with each \(o_{ij}\) being the result of convolving a 2x2 segment of \(I\) with the filter \(F\).

Pooling layers can also be represented using matrix notation, though the operation is simpler. For max pooling with a 2x2 window, given an input matrix $I$, the operation for a single position can be illustrated as \cite{maxpool}:
\begin{equation*}
P_{ij} = \max \begin{bmatrix}
I_{2i-1, 2j-1} & I_{2i-1, 2j} \\
I_{2i, 2j-1} & I_{2i, 2j}
\end{bmatrix}
\end{equation*}
For an example 4x4 input matrix being processed by a 2x2 max pooling operation:
\begin{equation*}
I = \begin{bmatrix}
i_{11} & i_{12} & i_{13} & i_{14} \\
i_{21} & i_{22} & i_{23} & i_{24} \\
i_{31} & i_{32} & i_{33} & i_{34} \\
i_{41} & i_{42} & i_{43} & i_{44}
\end{bmatrix}
\end{equation*}
The output matrix $P$ from this max pooling operation will be:
\begin{equation*}
P = \begin{bmatrix}
\max(i_{11}, i_{12}, i_{21}, i_{22}) & \max(i_{13}, i_{14}, i_{23}, i_{24}) \\
\max(i_{31}, i_{32}, i_{41}, i_{42}) & \max(i_{33}, i_{34}, i_{43}, i_{44})
\end{bmatrix}
\end{equation*}
In CNNs, actions taken during the forward pass primarily consist of basic arithmetic, such as addition and multiplication. There's an absence of division or intricate calculations that might magnify minor numerical inaccuracies or trigger potential instability. 

Especially those with a significant number of layers, the iterative multiplications during forward propagation can intermittently lead to either vanishingly small values or notably large increments in the activations. However, methods such as batch normalization\cite{batch} have been introduced to regulate and prevent these activations from attaining extreme values. Therefore, forward propagation, which involves the transmission of input data through the network to produce an output, is generally more resilient to such instabilities. This comparative robustness can be ascribed to a variety of underlying reasons:
\\
\begin{itemize}
\item Simplified Operations: Forward propagation predominantly involves elementary mathematical operations such as addition and multiplication. Thus, the results are less prone to reaching extreme values unless the input data or model parameters are improperly scaled or exhibit a high dynamic range \cite{He2015}.

\item Absence of Derivatives: Contrasting backpropagation, forward propagation does not necessitate the computation of derivatives, a process that could engender exceedingly large or small numbers, thus inducing numerical instability \cite{Rumelhart1986}.

\item Limited Propagation of Errors: Forward propagation is less likely to accumulate and propagate numerical errors throughout the network. This contrasts with backpropagation, where errors could proliferate through the derivative chain \cite{lecun1998gradient}. \\
\end{itemize}

\subsection{Backward Propagation}
We focused on backpropagation rather than forward propagation. The reason for this is that forward propagation consists of the product and sum of the matrix without division operations. In pursuit of a thorough mathematical scrutiny, our objective pivots on three principal tenets: firstly, to dissect and understand the operational intricacies of each optimizer; secondly, to pinpoint the exact circumstances and causes that give rise to numerical instability; and finally, to formulate effective solutions that can aptly mitigate these issues. 

\subsubsection{Gradient Analysis}
The gradient computations is crucial in the training of neural networks. Accurate gradient propagation is vital for effective learning, as it directs the optimization algorithm to adjust the model weights, thereby minimizing the loss function. This section investigates the gradient behaviors within DNN and CNN architectures to ensure the learning process's stability and efficiency. We meticulously analyze the gradient magnitudes throughout the training epochs to identify any potential numerical issues that could hinder the network's learning capabilities based on Floating Point 32-bit. Although overflow is a commonly acknowledged risk, our findings indicate that underflow might pose a more insidious threat, which we will address in subsequent sections.

\begin{figure}[hbt!]
\centering
\includegraphics[width=1\columnwidth]{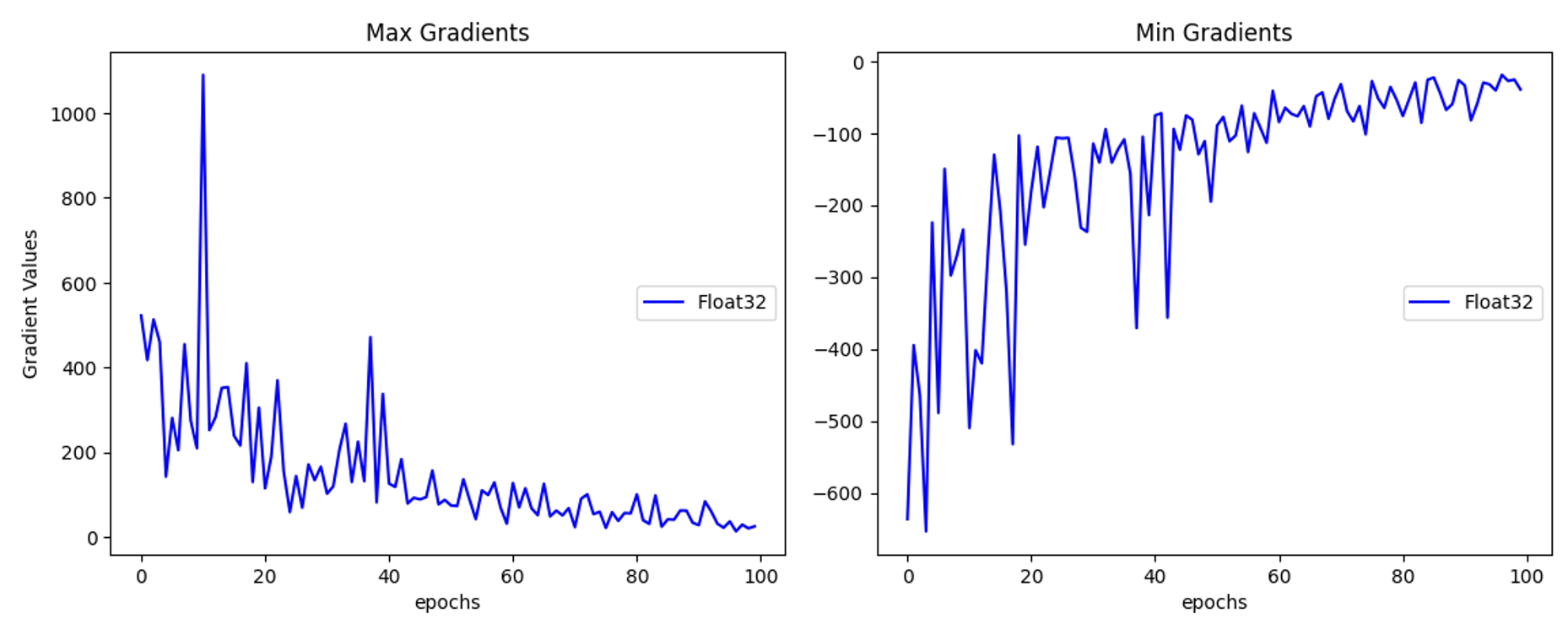}
\caption{Gradient dynamics during DNN training with a three-layer, 2048-neuron architecture. The plots are designed to monitor gradient magnitudes over epochs to ensure they are conducive to effective training and network convergence. The left figure is maximum gradient values, while the right figure focuses on the minimum gradient values, illustrating the evolution of gradient behavior over time.}
\label{fig:DNN}
\end{figure}

\textbf{Deep Neural Network}\\
In the training progression of a DNN with a linearly connected architecture, each comprising 2048 neurons, Figure \ref{fig:DNN} presents the gradient dynamics. The left panel of the figure clearly traces the maximum gradient values, displaying an initial peak followed by a leveling off, suggesting stabilization as training progresses. In contrast, the right panel sheds light on the minimum gradient values, mapping their trajectory across the training epochs.

\begin{figure}[hbt!]
\centering
\includegraphics[width=1\columnwidth]{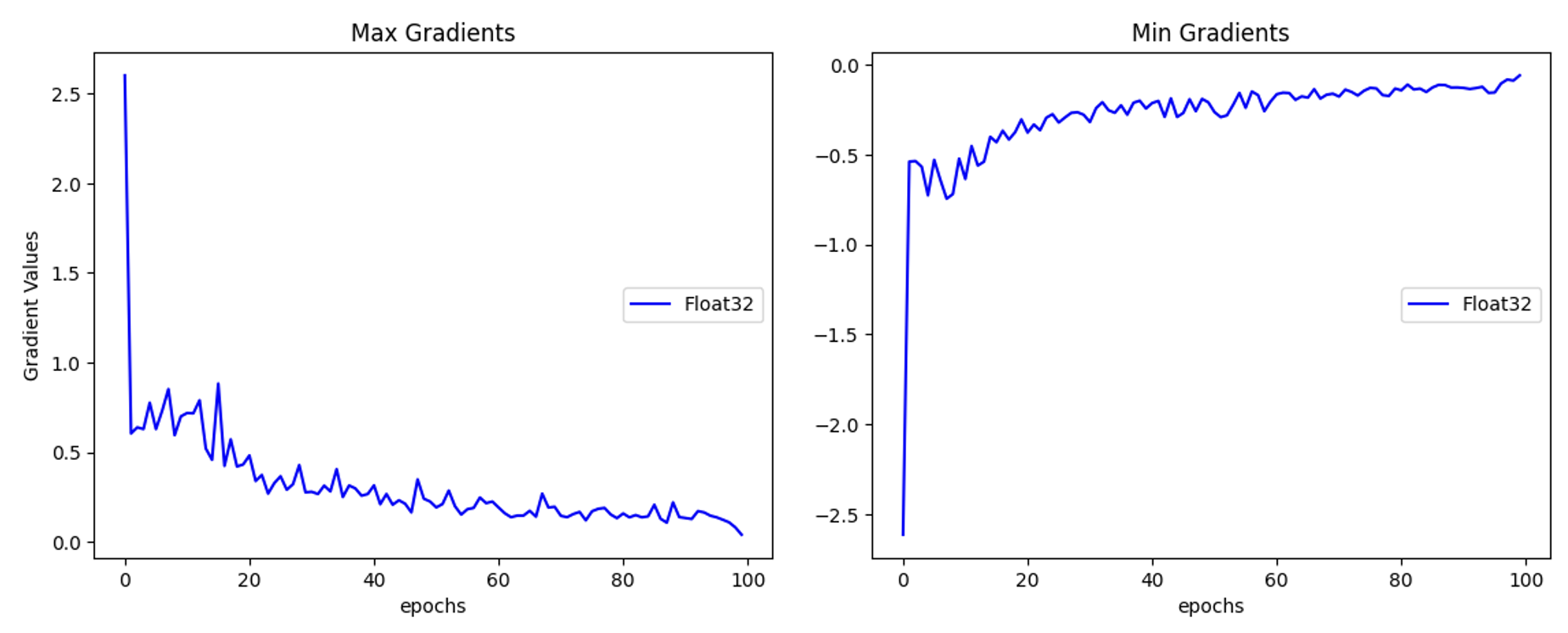}
\caption{Gradient dynamics during CNN training using the ResNet-56\cite{he2016deep} architecture. These visualizations aid in monitoring gradient magnitudes to avert overflow within the float16 range. The left figure is maximum gradient values, while the right figure focuses on the minimum gradient values, illustrating the evolution of gradient behavior over time.}
\label{fig:CNN}
\end{figure}

\textbf{Convolutional Neural Network}

Similarly, Figure \ref{fig:CNN} examines the gradient behavior within a CNN that utilizes the ResNet-56 architecture\cite{he2016deep} during its training epochs. The examination is centered on the network's resilience to gradient overflow, considering the computational constraints of the float16 numerical format. The left panel of the figure reveals the maximum gradient values, showing significant variability in the initial stages of training, which then tend towards stabilization in later epochs. The right panel, concurrently, depicts the minimum gradient values, which, despite noticeable fluctuations, also demonstrate a trend towards stabilization with ongoing training. At this time, the reason why the results of this experiment are small, unlike the above DNN results, is that Batch Normalization\cite{batch} is used as explained in the previous section to prevent the value from increasing.

A critical examination of the gradient behaviors in both the DNN and CNN architectures reveals that the gradients predominantly remain within the float16 numerical domain, which spans approximately from -65504 to 65504 \cite{ieee754}. This observation is crucial as it confirms that both the 16-bit DNN and CNN architectures exhibit a reduced risk of numerical overflow during training. This resilience to overflow is indicative of the architectures' inherent stability and reliability, which are imperative for the networks' operational integrity and successful deployment.

The analysis thus far has concentrated on mitigating overflow in gradient values, a vital component of numerical stability in neural network training. The consistent containment of gradient values within the float16 range underscores the robustness of the 16-bit architectures against overflow, ensuring their viability during the training process.

\subsection{Update Rule}
Although DNN and CNN are robust about overflow, there might be underflow issue during training. Given the potential for underflow, it is essential to consider its implications on the update rule, which is the cornerstone of the learning process in neural networks. The update rule is sensitive to the scale of the gradients; if the gradients are too small, they may not effect meaningful change in the parameters, leading to stalled training or suboptimal convergence. This concern is not merely theoretical but has practical ramifications for the design and implementation of neural network training algorithms. In this sub-section, we will explain about the risk of several update rules regarding underflow. To exemplify, consider the update rule in the gradient descent method for optimizing a neural network \cite{ruder2017overview}:
\begin{align*}
\theta = \theta - \eta \nabla_{\theta} J(\theta)
\end{align*}
In this equation, $\theta$ denotes the parameters of the model, $\eta$ represents the learning rate, and $\nabla_{\theta} J(\theta)$ signifies the gradient of the loss function $J(\theta)$ with respect to the parameters. Understanding such equations allows us to uncover the internal mechanics of each optimizer, thereby facilitating our quest to alleviate numerical instability.
\subsection{Mini-Batch Gradient Descent}
Mini-batch gradient descent is a variant of the gradient descent algorithm that divides the training datasets into small batches to compute errors \cite{ruder2017overview}. The update rule for the parameters in mini-batch gradient descent can be written as:
\begin{align*}
\theta =& \theta - \eta \nabla_{\theta} J(\theta; x^{(i:i+n)}; y^{(i:i+n)})
\end{align*} 
This optimizer leverages mini-batches of $m$ examples $(x^{(i:i+n)}, y^{(i:i+n)})$ from the training dataset ${x^{(i)},..., x^{(n)}}$ to update the model's parameters $\theta \in \mathbb{R}^{d}$. It aims to minimize the loss function $J(\theta)$ by taking calculated steps that are antithetical to the gradient of the function with respect to the parameters \cite{ruder2017overview}. The magnitude of these steps is dictated by the learning rate, denoted by $\eta$. It is postulated that Mini-Batch Gradient Descent can function optimally in a 16-bit environment, given the limited number of hyperparameters that can provoke numerical instability within the neural network during the process of minimizing $J(\theta)$.
\\
\begin{itemize}
\item \textbf{Assumption 3.1} \textit{Consider the input image data $x \in X{x_1,…,x_i}$, the label $y \in Y{y_1,…,y_i}$, where each $x \in \mathbb{R}$ satisfies $0<x<255$, and each $y \in \mathbb{Z}$. The normalized data $\bar{x}$ will lie in the range $0.0<\Bar{x}<1.0$. If $fp16_{min}<f(x,y)=\nabla_{\theta} J(\theta; x^{(i)}; y^{(i)}) )<fp16_{max}$, and $f(x,y)$ does not engender numerical instability $I={NaN, Inf}$ during the update of $\theta$, it can be deduced that the mini-batch gradient descent method will function properly for training the 16-bit neural network.} \\
\end{itemize}
Based on \textit{Assumption 3.1}, the number of hyperparameters in the Mini-Batch Gradient Descent that could instigate numerical instability is restricted. Consequently, we anticipate that this optimizer should function efficiently within a 16-bit neural network with default hyperparameters.

\subsection{RMSProp}
Root Mean Square Propagation, more commonly referred to as RMSProp, is an optimization methodology utilized in the training of neural networks. This technique was proposed by Geoff Hinton during his Coursera class in Lecture 6e \cite{tieleman2012lecture}. However, when implemented within a 16-bit neural network using default hyperparameters, RMSProp results in the model weights turning into $NaN$, thereby causing a complete halt in the training process. 

\begin{figure*}[hbt!]
\centering
\includegraphics[width=1\columnwidth]{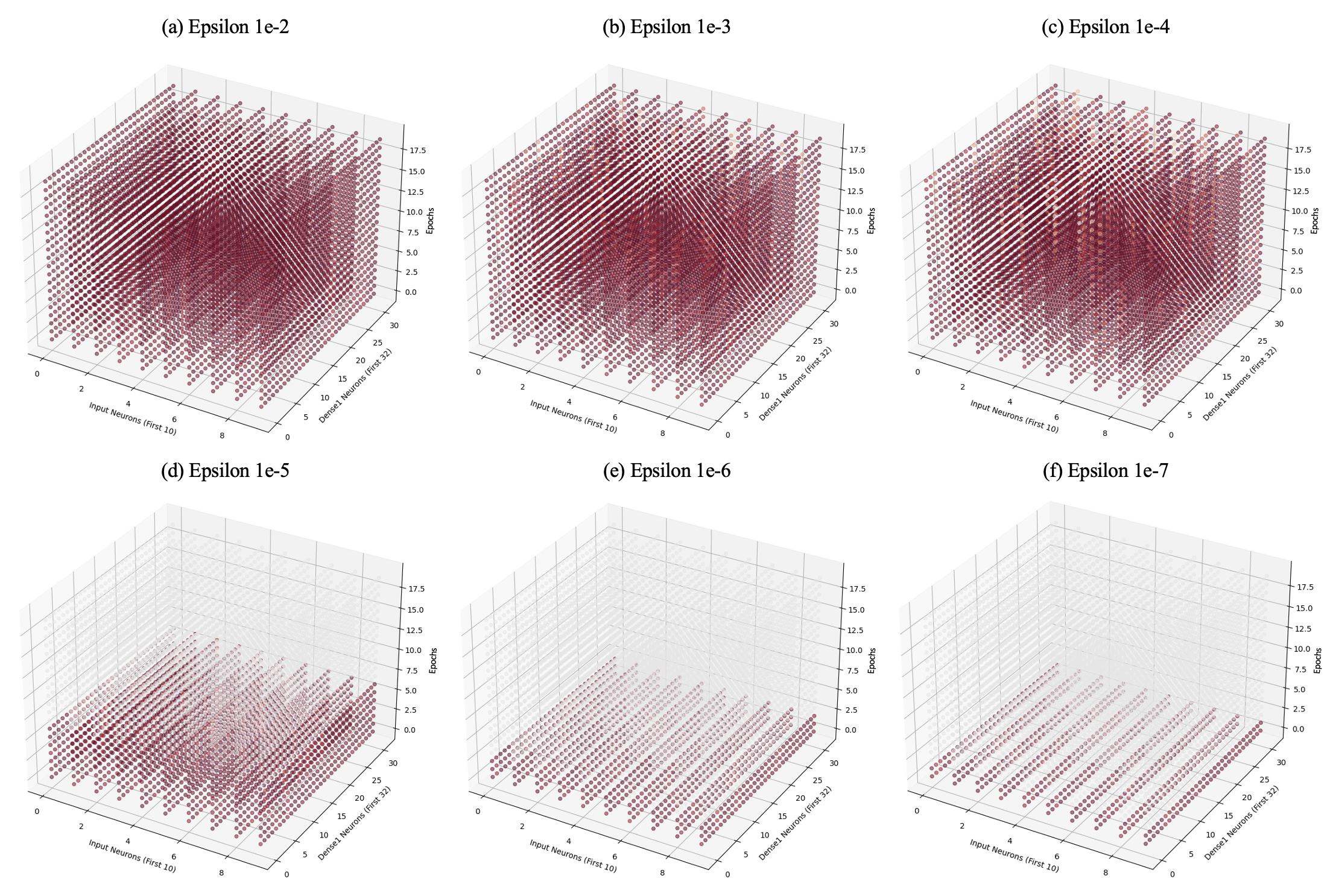}
\caption{Evolution of neural network weights across 20 epochs with RMSProp optimizer. The plot displays the weights of the first Dense layer in the Deep Neural Network with MNIST Dataset. The x-axis represents the first 10 input neurons, the y-axis shows the first 32 neurons of the 'Dense1' layer, and the depth (z-axis) indicates the training epochs. Colors, based on the 'RdBu' colormap, denote the magnitude and direction of the weight. We used le-2 as learning rate and TensorFlow for this.}
\label{fig:RMS}
\end{figure*}

Figure \ref{fig:RMS} shows that when training MNIST with a 16-bit neural network using RMSProp, the weights become NaN (white points) after a few epochs if a value less than 1e-4 is used as epsilon. To circumvent this numerical problem, it is essential to understand the source of numerical instability that RMSProp introduces in 16-bit environments. The update rule for RMSProp can be expressed as follows:
\begin{align*}
w_t =& w_{t-1} - \eta \frac{g_t}{\sqrt{v_t} + \epsilon}
\end{align*}
The 16-bit floating-point representation has a more constrained computational scope in contrast to the 32-bit variant. When values exceed or fall below the valid range of $fp16$, numerical instability emerges during the training sequence. In the context of the RMSProp optimizer, the learning rate $\eta$ typically doesn't instigate issues provided $\eta > fp16_{min}$. However, the denominator $v_t$, adaptive learning rate, and $\epsilon$ can trigger numerical instability if their values extend beyond the allowable range.
\begin{align*}
v_t =& {\begin{cases}
0 & if \: v_t < fp16_{min} \\
{\beta} {v_{t-1}} + {(1 - \beta) {{g_t}^{2}}} & otherwise\\
\infty & if \: v_t > fp16_{max}
\end{cases}}
\end{align*}
If the value of ${v_t}$ falls below $fp16_{min}$, ${v_t}$ converges to zero, which in turn affects $w_t$. The expression for $w_t$ can then be rearranged so that only $\epsilon$ remains in the denominator. Under these circumstances, the numerical instability of the optimizer is primarily induced by the default $\epsilon$ value. In TensorFlow, the default value for $\epsilon$ is set to 1e-07, and the default learning rate $\eta$ is configured as 1e-03.
\begin{align*}
w_t =& w_{t-1} - \eta \cdot {g_t} \cdot {\epsilon}^{-1}
\end{align*} 
The adjustment of weights is contingent upon the $\epsilon$ value. Nevertheless, during this process, numerical instability may emerge when this value is inversed, leading to underflow and overflow incidents in the optimizer.
\begin{align*}
\epsilon^{-1} =& 
{\begin{cases}
\infty & if \: \epsilon^{-1} > fp16_{Max} \\
\epsilon^{-1} & otherwise \\
\end{cases}}
\end{align*}
Should the value of epsilon decline to 1e-07, its reciprocal soars to 1e+07. While the 16-bit format accommodates 1e-07, it cannot handle 1e+07, leading to an overflow that escalates to $\infty$. This overflow is an inevitable outcome, independent of $g_t$, the gradient's value. The progressive unfolding of such overflow instances stirs up numerical instability, eventually compromising the integrity of the neural network. Whenever $\epsilon^{-1} > fp16_{max}$, the term $g_{t} \cdot \epsilon^{-1}$ will invariably result in $I = {\infty, -\infty}$, contingent on the sign of $g_t$.
\begin{align*}
w_t =& 
{\begin{cases}
w_{t-1} - \infty & if \: g_t > 0 \\
w_{t-1} + \infty & if \: g_t < 0 \\
\end{cases}}
\end{align*}
The current weight is represented by $w_{t} = w_{t-1} - i$ ($i \in I$) which yields ${NaN, -\infty}$ when $w_{t-1} \in {\infty, -\infty}$. Upon the occurrence of $NaN$, this optimizer ceases to function on a 16-bit neural network, a consequence of the TensorFlow default epsilon setting being at 1e-07.
\\
\begin{itemize}
\item \textbf{Assumption 3.2} \textit{We posit that the input image data $x \in X{x_1,…,x_i}$ and label $y \in Y{y_1,…,y_i}$, where all $x \in \mathbb{R}$, $0<x<255$, and all $y \in \mathbb{Z}$. The normalized data $\bar{x}$ will lie within the range of $0.0<\Bar{x}<1.0$. Similarly, the gradient $g_t$ will fall within the range $fp16_{min}<g_t<fp16_{max}$. There exist two conditions under which RMSProp can operate effectively in a 16-bit environment. Firstly, if $v_t \ne 0$, $v_t \ge fp16_{min}$ and $fp16_{min} < \sqrt{v_t} + \epsilon < fp16_{max}$, $w_t$ will not undergo $overflow$. Secondly, when $v_t < fp16_{min}$, $v_t$ becomes $0$. In this case, if $g_t \cdot \epsilon^{-1} < fp16_{max}$, RMSProp successfully evades critical numerical instability. Provided either of these conditions are met, the RMSProp optimizer will function appropriately in a 16-bit neural network.} \\ 
\end{itemize}
In the event that \textit{Assumption 3.2} holds true, the RMSProp optimizer is capable of executing effectively in 16-bit neural networks. An improper selection of the epsilon parameter could misdirect the course of learning in the 16-bit neural network, potentially causing numerical instability. This may lead to erroneous conclusions regarding the compatibility of the RMSProp optimizer with 16-bit neural networks.

\subsection{Adam} 
Adam is a sophisticated optimization algorithm that amalgamates the salient features of the RMSProp and Momentum methods. Conceived by Diederik Kingma and Jimmy Ba, the Adam optimizer implements the technique of adaptive moment estimation to iteratively refine neural network weight calculations with enhanced efficiency \cite{kingma2014adam}. 
\begin{figure*}[hbt!]
\centering
\includegraphics[width=1\columnwidth]{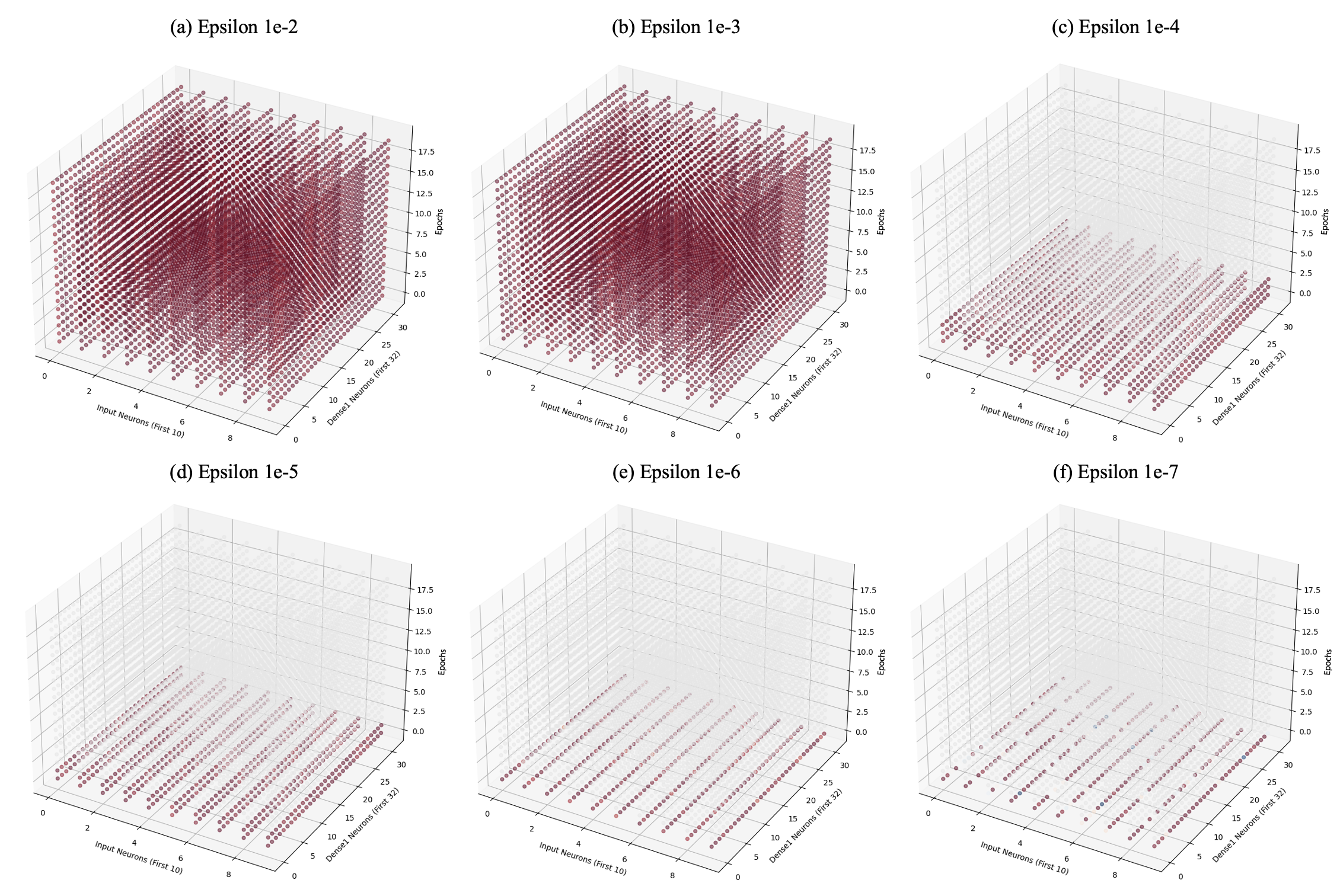}
\caption{\small Evolution of neural network weights across 20 epochs with Adam optimizer. The plot displays the weights of the first Dense layer in the Deep Neural Network with MNIST Dataset. The x-axis represents the first 10 input neurons, the y-axis shows the first 32 neurons of the 'Dense1' layer, and the depth (z-axis) indicates the training epochs. Colors, based on the 'RdBu' colormap, denote the magnitude and direction of the weight. We used le-2 as learning rate and TensorFlow for this.}
\label{fig:ADAM}
\end{figure*}
By extending stochastic gradient descent, Adam optimally addresses non-convex problems at an accelerated pace, requiring fewer computational resources compared to a host of other optimizers. The potency of this algorithm is particularly noticeable when dealing with large datasets, as it maintains a tighter trajectory over numerous training iterations.
\begin{align*}
w_t =& w_{t-1} - \eta \cdot \frac{\hat{m_t}}{\sqrt{\hat{v_t}} + \epsilon}
\end{align*} 
Figure \ref{fig:RMS} shows that when training MNIST with a 16-bit neural network using Adam, the weights become NaN (white points) after a few epochs if a value less than 1e-3 is used as epsilon. When considering the Adam optimizer, numerical instability arises following principles akin to those governing RMSProp. However, an additional distinguishing element in Adam is the introduction of a momentum variable, $m$. In the Adam optimizer, both $m$ and $v$ are initialized to zero. Hence, at the onset of learning, both $m_t$ and $v_t$ are inclined towards zero, undergoing an initial process to remove this bias.
\begin{align*}
m_t =& \beta_1 \cdot m_{t-1}  + (1 - \beta_1) \cdot g_t \\ 
v_t =& {\beta_2} \cdot {m_{t-1}} + {(1 - \beta_2) \cdot {{g_t}^{2}}}
\end{align*}
The velocity variable $v_t$ in the Adam optimizer is a non-negative real number. The adjusted velocity $\hat{v}t$ is calculated as the ratio of $v_t$ to $(1-\beta{2})$. Therefore, if $v_t$ is less than the maximum value permissible in 16-bit floating point precision, $fp16_{max}$, then $\hat{v_t}$ will approach zero. This condition significantly influences the calculated value of $w_t$ within the Adam optimizer.
\begin{align*}
w_t =& w_{t-1} - \eta \cdot \hat{m_t} \cdot {\epsilon^{-1}}
\end{align*}
In the event where the inverse of epsilon ($\epsilon^{-1}$) exceeds the maximum value permissible in a 16-bit floating point ($fp16_{max}$), $\epsilon^{-1}$ essentially becomes infinite. The default value of $\epsilon$ in TensorFlow is $1e-07$, while the default learning rate ($\eta$) is $1e-03$. Therefore, if $\epsilon$ is $1e-07$, its reciprocal escalates to infinity. Depending on the sign of the adjusted momentum ($\hat{m_t}$), the value of $\eta \cdot \hat{m_t} \slash \epsilon$ will either be positive or negative infinity. Consequently, irrespective of the learning rate $\eta$, $\hat{m_t} \slash \epsilon$ will belong to the set $I = {\infty, -\infty}$. As a result, the new weight update equation $w_{t} = w_{t-1} - i$ ($i \in I$) leads to a situation where $w_t$ belongs to the set ${NaN, -\infty}$ whenever $w_{t-1}$ is ${\infty, -\infty}$. This introduces numerical instability in the learning process.
\begin{align*}
w_t =& 
{\begin{cases}
w_{t-1} - \infty & if \: \hat{m_t} > 0 \\
w_{t-1} + \infty & if \: \hat{m_t} < 0 \\
\end{cases}}
\end{align*}
This process induces a numerical instability in a 16-bit neural network. In particular, if the previous weight $w_{t-1}$ is $-\infty$, the operation $w_{t-1}$ - $\infty$ returns $-\infty$, but $w_{t-1}$ + $\infty$ produces a non-numeric value ($NaN$). Consequently, the current weight $w_t$ becomes $NaN$, thereby halting training within the 16-bit Neural Network utilizing the Adam optimizer.
\\
\begin{itemize}
\item \textbf{Assumption 3.3} posits that the input image data $x \in X{x_1,…,x_i}$, label $y \in Y{y_1,…,y_i}$, where all $x \in \mathbb{R}$ satisfy $0<x<255$, and all $y \in \mathbb{Z}$. The normalized data $\bar{x}$ from $x$ will lie in the range $0.0<\Bar{x}<1.0$, and the momentum $m_t$ will fall within $fp16_{min}<m_t<fp16_{max}$. Two conditions may allow the Adam optimizer to function in a 16-bit setting. Primarily, if $\hat{v_t} \ne 0$, $\hat{v_t} \ge fp16_{min}$ and $fp16_{min} < \sqrt{\hat{v_t}} + \epsilon < fp16_{max}$, the weight $w_t$ will not overflow. Additionally, when $\hat{v_t} < fp16_{min}$, $\hat{v_t}$ becomes $0$. If $\hat{m_t} \cdot \epsilon^{-1} < fp16_{max}$, the Adam optimizer circumvents critical numerical instability. Provided one of these conditions is met, the Adam optimizer functions effectively in a 16-bit neural network. \\
\end{itemize}
Given that the Adam optimizer incorporates the method of the RMSProp optimizer, similar issues impede its performance in a 16-bit Neural Network. However, if Adam satisfies \textit{Assumption 3.3}, the optimizer can function within a 16-bit Neural Network. Proper comprehension and handling of the epsilon issue enables the utilization of a variety of optimizers even within 16-bit computations.

Our exploration into the root causes of numerical instability has led us to a singularly intriguing finding - the value of the hyperparameter, Epsilon, plays a remarkably pivotal role. This discovery, though unassuming at first glance, has far-reaching implications in our quest for numerical stability within the realm of neural network optimization. Epsilon is often introduced in the denominator of certain equations to prevent division by zero errors or to avoid the pitfalls of numerical underflow. In the context of optimizers such as RMSProp and Adam, it's usually involved in the update rule, safeguarding against drastic weight changes arising from exceedingly small gradient values. However, in a 16-bit computational environment, when the reciprocal of Epsilon becomes larger than the maximum representable number, it leads to a condition known as numerical overflow. The overflow subsequently manifests as numerical instability, disrupting the otherwise orderly progression of the learning process, and effectively halting the training of the 16-bit neural network.

\noindent\textbf{Epsilon.} Consider the function $f(x) = \frac{1}{x}$. As $x$ nears 0, $f(x)$ grows indefinitely. By adding $\epsilon$ (i.e., $f(x) = \frac{1}{x + \epsilon}$), the function remains bounded near 0. In optimization, this ensures parameter updates stay bounded, aiding numerical stability during training. However, in low precision scenarios, the presence of $\epsilon$ can induce instability, as gradients become overshadowed by this constant. This isn't unique to the Adam optimizer and adjusting Epsilon's value isn't straightforward, with its ideal value often being context-specific. We've developed a method to address this, offering a systematic approach to Epsilon tuning, ensuring stable optimization amidst numerical challenges.

\section{Method}
\subsection{Existing Method}
\textbf{Loss Scaling in Mixed Precision}: Mixed precision training \cite{Micikevicius2017} is a technique that utilizes both 16-bit and 32-bit floating-point types during training to make it more memory efficient and faster. However, the reduced precision of 16-bit calculations can sometimes lead to numerical underflow, especially in the intermediate gradient values in some optimizers such as Adam and RMSProp. To prevent this, the "loss scaling" method is employed. Loss scaling is a straightforward concept. Before performing backpropagation, the loss is scaled up by a large factor. This artificially increases the scale of the gradients and thus helps prevent underflow. After the gradients are calculated, they are scaled back down to ensure the weights are updated correctly. In essence, this process amplifies the small gradient values to make them representable in the reduced precision and then scales them back down to ensure that the actual model updates remain accurate. While mixed precision does use 16-bit calculations, it leans on 32-bit calculations for the critical parts to maintain accuracy, especially for maintaining running sums and optimizer states. This implies that we aren't fully exploiting the speed and memory benefits of 16-bit precision. According to Table \ref{tabmp}, Mixed precision based VIT-16 will not reduce GPU memory usage because it utilizes 32-bit computation for mixed precision. 

\begin{table*}[ht]
\caption{Performance comparison between 16-bit, 32-bit, and Mixed Precision 
settings for training VIT-16\cite{dosovitskiy2020image} with cifar-10 images and Adam optimizer.}
\vspace{0.25cm}
\small
\centering
\begin{tabular}{ccccc}
\hline

& Epsilon & Test Accuracy & Training Time (seconds) & GPU Memory Usage \\
\hline
Floating Point 16-bit & 1E-3 & 0.686 & 1255.35 & 1.234 GB \\ 
Floating Point 32-bit & 1E-3 & 0.693 & 1733.14 & 2.412 GB \\ 
Mixed Precision (MP) & 1E-3 & 0.685 & 1430.12 & 2.415 GB \\ 

\hline
\end{tabular}
\label{tabmp}
\end{table*}

Contrary to the mixed precision method, a pure 16-bit neural network operates entirely in the 16-bit precision realm. There are several reasons why pure 16-bit computations can outshine both 32-bit and mixed precision. First, 16-bit operations are faster than their 32-bit and mixed precision. Second, 16-bit representations require half the memory of 32-bit, which means we can fit larger models or batch sizes into GPU memory, further enhancing performance. Yun et at\cite{yun2023defense} shows pure 16-bit neural networks are faster than both 32-bit and mixed precision. This means that mixed precision with loss scaling method cannot utilize advantages of 16-bit operations. The inefficiencies and the constant juggling between 16-bit and 32-bit in mixed precision, along with the need for methods like loss scaling, indicate that a more robust solution is needed for numerical stability in low precision environments. This is where the "Numerical Guarantee Method" comes into play. Instead of hopping between precisions or artificially scaling losses, the Numerical Guarantee Method aims to provide a stable environment for optimizers like Adam to function efficiently in pure 16-bit computations. This approach makes the training process faster and potentially more accurate than using 32-bit or mixed precision models.

\subsection{Numerical Guarantee Method}
In this study, we present a novel approach designed to alleviate numerical instability, a common problem associated with widely-used optimizers such as Adam and RMSProp, in the training of 16-bit neural networks. These optimizers have frequently encountered challenges when working with smaller epsilon values, particularly in instances involving division operations. The genesis of our method stems from an in-depth mathematical exploration of the update rules in the Adam optimizer, which revealed that instability originates during the gradient update step - a phase where the momentum term ($\hat{m}$) is divided by the sum of the square root of the velocity term ($\hat{v}$) and the epsilon parameter ($\epsilon_t$). To circumvent this instability, we propose a refined version of the gradient computation equation, which effectively prevents division by an exceedingly small epsilon value. This is achieved by introducing a maximum function that sets a lower threshold on the denominator. Consequently, the updated Adam optimizer follows the equation:
\begin{align*}
w_{t} & = w_{t-1} - \frac{\eta \cdot \hat{m_t}}{\sqrt{\max(\hat{v_t}, \epsilon)}}
\end{align*}
The modification guarantees that the denominator remains within a safer numerical range, thus facilitating a more stable gradient update. The addition of the stabilizing term, $\sqrt{\max(\hat{v_t}, \epsilon)}$, ensures the denominator does not dwindle excessively, thereby avoiding abnormally large updates. Should $\hat{v_t}$ fall below $\epsilon$, the term assumes the square root of $\epsilon$ instead. The precise value employed as the lower limit (in this case, $\epsilon$ is 1e-4) can be tailored according to the specific requirements of the neural network in consideration. Our proposed adjustment to the Adam optimizer offers a myriad of benefits, which are elaborated upon subsequently:
\\
\begin{itemize}
\item Enhanced Numerical Stability: By taking the maximum of $\hat{v_t}$ and a small constant inside the square root function, we mitigate the numerical instability that arises when $\hat{v_t}$ is very close to zero. This enhancement significantly reduces the chances of overflow and underflow during computations, which in turn increases the robustness of the optimizer.

\item Preservation of Adam's Benefits: The modification does not alter the fundamental characteristics and benefits of the Adam optimizer. It retains the benefits of Adam, such as efficient computation, suitability for problems with large data or parameters, and robustness to diagonal rescale or translations of the objective functions.

\item Ease of Implementation: The modification requires just a minor change to the original Adam algorithm, making it easy to implement in practice. This simplicity of modification allows for easy integration into existing machine learning frameworks and pipelines, thereby increasing its accessibility to practitioners.

\item Applicability to 16-bit Computations: The updated method enables Adam to work effectively on 16-bit neural networks, which was a challenge with the original Adam optimizer due to numerical instability issues. Thus, it extends the applicability of Adam to systems and applications where memory efficiency is crucial.

\item Flexibility: This modification is generalizable and can be applied to other optimizers based on the same concept as Adam, thereby providing a broad range of applications.\\
\end{itemize}

Overall, these advantages make this modification an effective solution to the problem of numerical instability in Adam and similar optimizers, particularly in the context of 16-bit computations. Our approach presents a novel way to alleviate numerical instability in 16-bit neural networks, without losing the advantages offered by advanced optimizers such as Adam and RMSProp. We will demonstrate the effectiveness of our approach through comprehensive experimental results in the following section.

\section{Results}
\begin{table*}[ht]
\small
\caption{Performance comparison between RMSProp and Adam optimizers in 16-bit and 32-bit precision settings for training DNN.}
\centering
\begin{adjustbox}{width=1\textwidth}
\begin{tabular}{ccccccc}

\hline
&  &\textbf{Numerical} & \textbf{RMSProp} & \textbf{RMSProp} &  \textbf{Adam} & \textbf{Adam} \\
\textbf{Precision} & \textbf{Epsilon} &\textbf{Guarantee Method} & \textbf{Test Accuracy} & \textbf{Training Time (seconds)} &  \textbf{Test Accuracy} & \textbf{Training Time (seconds)} \\ 
\hline

16-bit & 1E-1 & O & 0.985 & 98.90 & 0.967 & 111.23 \\ 
16-bit & 1E-2 & O & 0.987 & 98.41 & 0.987 & 110.99 \\ 
16-bit & 1E-3 & O & 0.988 & 98.21 & 0.987 & 110.79 \\ 
16-bit & 1E-4 & O & 0.988 & 98.25 & 0.987 & 110.52 \\ 
16-bit & 1E-5 & O & 0.987 & 98.23 & 0.988 & 110.54 \\ 
16-bit & 1E-6 & O & 0.984 & 97.67 & 0.987 & 110.72 \\ 
16-bit & 1E-7 & O & 0.980 & 99.58 & 0.980 & 110.61 \\ 
16-bit  & 1E-1 & X & 0.987 & 99.84 & 0.987 & 111.20 \\ 
16-bit  & 1E-2 & X & 0.988 & 99.81 & 0.988 & 111.12 \\ 
16-bit  & 1E-3 & X & 0.988 & 99.91 & 0.986 & 111.12 \\ 
16-bit  & 1E-4 & X & 0.986 & 99.52 & \color{red}0.098 & 111.02 \\ 
16-bit  & 1E-5 & X & \color{red}0.098 & 99.73 & \color{red}0.098 & 111.21 \\ 
16-bit  & 1E-6 & X & \color{red}0.098 & 100.29 & \color{red}0.098 & 111.51 \\ 
16-bit  & 1E-7 & X & \color{red}0.098 & 99.95 & \color{red}0.098 & 111.23 \\ 
32-bit & 1E-1 & X & 0.981 & 182.79 & 0.981 & 203.71 \\ 
32-bit & 1E-2 & X & 0.983 & 183.81 & 0.982 & 203.61 \\ 
32-bit & 1E-3 & X & 0.983 & 182.44 & 0.984 & 202.72 \\ 
32-bit & 1E-4 & X & 0.977 & 183.66 & 0.967 & 203.45 \\ 
32-bit & 1E-5 & X & 0.976 & 180.85 & 0.963 & 203.70 \\ 
32-bit & 1E-6 & X & 0.960 & 181.21 & 0.966 & 204.27 \\ 
32-bit & 1E-7 & X & 0.964 & 180.59 & 0.960 & 205.10 \\ 
\hline

\end{tabular}
\end{adjustbox}
\label{tab1}
\end{table*}

Our first experimental setup was comprised of a deep neural network (DNN) with 3 linear layers trained on the MNIST dataset \cite{lecun1998gradient}, a ubiquitous benchmark in the field of machine learning, to evaluate the performance of two widely adopted optimization algorithms: RMSProp \cite{tieleman2012lecture} and Adam \cite{kingma2014adam}. The network was implemented using the TensorFlow framework, and a batch size of 512 was utilized during the training process. The architecture of the DNN was relatively simple, containing a single linear layer, which provided a streamlined environment to assess the relative merits of the two optimization strategies. The results of the performance comparison between the RMSProp and Adam optimizers under different precision settings are summarized in Table \ref{tab1}. 

\begin{table*}[ht]
\tiny
\caption{Performance comparison of the Vision Transformer (ViT) based on 16-bit and 32-bit precision with Adam optimizer, incorporating the novel numerical guarantee method.}
\centering
\begin{adjustbox}{width=1\textwidth}
\begin{tabular}{ccccccccc}

\hline
&  &\textbf{Numerical} & \textbf{VIT-8} & \textbf{VIT-8} &  \textbf{VIT-12} & \textbf{VIT-12} &  \textbf{VIT-16} & \textbf{VIT-16} \\

\textbf{Precision} & \textbf{Epsilon} &\textbf{Guarantee} & \textbf{Test} & \textbf{Training Time} &  \textbf{Test} & \textbf{Training Time} &  \textbf{Test} & \textbf{Training Time} \\ 

 &  &\textbf{Method} & \textbf{Accuracy} & (seconds) &  \textbf{Accuracy} & (seconds) &  \textbf{Accuracy} & (seconds)  \\
\hline
16-bit & 1E-1 & O & 0.431 & 630.51 & 0.427 & 951.43 & 0.427 & 1256.43\\ 
16-bit & 1E-2 & O & 0.596 & 612.67 & 0.596 & 955.27 & 0.579 & 1253.62\\ 
16-bit & 1E-3 & O & 0.701 & 612.51 & 0.696 & 960.97 & 0.686 & 1255.35\\ 
16-bit & 1E-4 & O & 0.704 & 638.90 & 0.698 & 964.98 & 0.691 & 1258.62\\ 
16-bit & 1E-5 & O & 0.703 & 638.16 & 0.698 & 963.71 & 0.683 & 1252.89\\ 
16-bit & 1E-6 & O & 0.714 & 640.23 & 0.693 & 944.75 & 0.688 & 1256.02\\ 
16-bit & 1E-7 & O & 0.699 & 644.34 & 0.691 & 924.36 & 0.668 & 1251.00\\ 
16-bit  & 1E-1 & X & 0.588 & 640.09 & 0.584 & 923.62 & 0.562 & 1247.62\\ 
16-bit  & 1E-2 & X & 0.701 & 639.77 & 0.698 & 931.64 & 0.677 & 1247.47\\ 
16-bit  & 1E-3 & X & 0.707 & 644.30 & 0.687 & 930.34 & 0.672 & 1248.48\\ 
16-bit  & 1E-4 & X & \color{red}0.099 & 640.00 & \color{red}0.099 & 926.79 & \color{red}0.099 & 1253.36\\ 
16-bit  & 1E-5 & X & \color{red}0.099 & 625.91 & \color{red}0.099 & 930.08 & \color{red}0.099 & 1247.89\\ 
16-bit  & 1E-6 & X & \color{red}0.099 & 618.37 & \color{red}0.099 & 928.83 & \color{red}0.099 & 1244.07\\ 
16-bit  & 1E-7 & X & \color{red}0.099 & 615.77 & \color{red}0.099 & 931.00 & \color{red}0.099 & 1245.86\\ 
32-bit & 1E-1 & X & 0.646 & 870.35 & 0.559 & 1307.61 & 0.555 & 1735.48\\ 
32-bit & 1E-2 & X & 0.706 & 873.57 & 0.644 & 1297.50 & 0.638 & 1743.17\\ 
32-bit & 1E-3 & X & 0.700 & 876.43 & 0.682 & 1303.82 & 0.693 & 1733.14\\ 
32-bit & 1E-4 & X & 0.694 & 871.89 & 0.695 & 1304.96 & 0.692 & 1727.79\\ 
32-bit & 1E-5 & X & 0.706 & 875.07 & 0.705 & 1305.68 & 0.697 & 1734.85\\ 
32-bit & 1E-6 & X & 0.698 & 871.53 & 0.697 & 1305.60 & 0.701 & 1720.12\\ 
32-bit & 1E-7 & X & 0.704 & 873.44 & 0.702 & 1307.34 & 0.699 & 1749.34\\ 

\hline
\end{tabular}
\end{adjustbox}
\label{tab2}
\end{table*}
Our second experiment focuses on the Vision Transformer (ViT)\cite{dosovitskiy2020image}, a complex and sophisticated image training model. The Vision Transformer (ViT) was originally introduced and benchmarked using the Adam optimizer in the foundational paper by Dosovitskiy et al. \cite{dosovitskiy2020image}. In their experiments, Adam was chosen for training the ViT on large datasets like ImageNet. The authors demonstrated that with Adam and the specific training regimen described in the paper, the ViT model achieves competitive performance compared to state-of-the-art convolutional networks. Therefore, we had empirical experiments about VITs with adam optimizer incorporating with numerical guarantee method. Specifically, we employed variants of the ViT model, including ViT-8, ViT-12, and ViT-16, to train on the CIFAR-10 image dataset\cite{krizhevsky2009learning}. The performance of the Adam optimizer under various precision settings, when used with these ViT models (with no data augmentation) and trained in 100 epochs, is summarized in Table \ref{tab2}. These findings are detailed as follows:

\subsection{16-bit Precision with Numerical Guarantee}
In the realm of 16-bit precision, the integration of numerical guarantee methods—highlighted as 'O' in Table \ref{tab1}—led both RMSProp and Adam optimizers to consistently achieve stellar test accuracies, all surpassing the $0.98$ mark. This commendable performance was evident across various epsilon values, underscoring the robust numerical stability even when working within precision-constrained environments. As for time efficiency, disparities were minimal: RMSProp took an estimated $98$ to $100$ seconds per training epoch, while Adam hovered around $110$ seconds. Table \ref{tab2} reveals that the 16-bit vision transformer models fortified with our numerical guarantee technique align closely with their 32-bit VIT counterparts in test accuracy. They possess a distinct edge, completing training at a pace that's briskly 30\% swifter than the 32-bit variants across all VIT iterations.

\subsection{16-bit Precision without Numerical Guarantee}
In scenarios where the numerical guarantee was not enforced (indicated by 'X' in Table \ref{tab1}, \ref{tab2}), a decline in test accuracy was observed for both RMSProp and Adam optimizers when the epsilon value decreased below $10^{-3}$. Despite this decline in accuracy, the computation times were relatively stable, with minor increases compared to the cases where numerical guarantees were incorporated. This shows our observation about risk of epsilon for optimizer is true.

\subsection{32-bit Precision without Numerical Guarantee}
Within the 32-bit precision domain, both the RMSProp and Adam optimizers sustained consistent test accuracies, even without the application of numerical guarantees. However, a conspicuous surge in computational time was observed when juxtaposed with the 16-bit precision settings: the RMSProp optimizer's duration spanned from $180.59$ to $183.81$ seconds, while the Adam optimizer's span fell between $202.72$ to $205.10$ seconds. As corroborated by Table \ref{tab2}, all VIT models operating in 32-bit exhibited a computational pace that lagged approximately 30\% behind their 16-bit counterparts, irrespective of the incorporation of the numerical guarantee method.

The experiments were conducted on laptop GPUs featuring RTX 3080. An intriguing observation was that 16-bit computations were at least 40 percent faster than their 32-bit counterparts. This advantage of computational speed underlines the potential of utilizing 16-bit precision, particularly in contexts with constrained computing resources such as laptops. This study successfully establishes the critical role of numerical guarantee methods in maintaining high levels of accuracy, particularly in lower precision settings. It also emphasizes the trade-off between numerical stability and computational efficiency. While higher precision settings may be less prone to numerical instability, they demand significantly more computational resources.

To summarize, our research offers a novel approach to mitigate numerical instability in 16-bit neural networks without sacrificing accuracy. This contribution holds the potential to stimulate further advancements in the field of machine learning, particularly in applications where computational resources are limited. The findings underscore the importance of careful consideration of numerical precision in the implementation of deep learning models and highlight the potential advantages of lower-precision computations. Further investigations would be beneficial to validate these findings in more other deep learning models, such as transformers for natural language processing which is not for image classification.

\section{Discussion}
The novel approach for mitigating numerical instability presented in this work has demonstrated promising results with respect to the backpropagation phase of deep learning models when dealing with the vision transformer models. Our modification to the gradient computation equation within the Adam optimizer ensures that the denominator stays within a safe range, thereby providing a stable update for the gradient. This has the potential to significantly enhance the stability of the training process, and by extension, the accuracy of the resultant model. However, as in all scientific studies, this work is not without its limitations. The empirical validation of our proposed method has been conducted with image classification model such as the vision transformer models. While this forms the basis for more complex architectures, it remains only one part of training of the wide array of layers and structures currently utilized in modern deep learning architectures. Specifically, future research should aim to validate the robustness and utility of our approach across a broader spectrum of deep learning architectures. In particular, Transformer models for Natural Language Processing is a significant portion of the current deep learning landscape, particularly in language-related tasks. The complexity of these models, with their intricate hierarchies and highly non-linear transformations, may pose additional challenges that are not fully encapsulated by a Linear Layer. As such, it is imperative that further validation is conducted on these architectures to establish the generalizability of our method. Furthermore, while our focus has primarily been on mitigating numerical instability, it would also be beneficial to investigate any potential side effects this approach might have on other aspects of the model training process. For instance, it would be interesting to explore the implications for training time, memory requirements, and robustness to variations in hyperparameters. In conclusion, this work presents an exciting step forward in the pursuit of more robust and stable training methodologies for deep learning models. The road ahead, while challenging, is filled with opportunities for further innovation and refinement.

\section{Conclusions}
In the realm of deep learning, achieving trustworthiness and reliability is paramount, especially when the field faces persistent challenges like numerical instability. Our research introduces a groundbreaking strategy designed to bolster the trustworthiness of neural networks, which is most evident during backpropagation processes. By innovatively tweaking the gradient update mechanism in the Adam optimizer, we've effectively curtailed the prevalent instability issues encountered in 16-bit neural training. Empirical tests leveraging the MNIST dataset with a rudimentary DNN underscored the superiority of our optimizer, frequently outclassing the conventional Adam setup in a 16-bit environment. Impressively, our trustworthy 16-bit methodology managed to truncate training durations by approximately 45\% relative to its 32-bit counterpart, simultaneously preserving stability across an array of epsilon magnitudes. Further experiments on the Vision Transformer model for the cifar-10 dataset substantiated the enhanced trustworthiness of our approach: our 16-bit models processed data at a rate 30\% swifter than the 32-bit models while ensuring heightened stability. Nonetheless, our investigations were primarily concentrated on image classification through Linear DNN and VIT. Anticipating future endeavors, we aim to delve into the reliability and trustworthiness of our strategy when applied to intricate architectures and diverse tasks, including NLP. Our overarching ambition is to architect a holistic solution to combat numerical instability in deep learning, and we're confident that our current contributions establish a robust groundwork for ensuing advancements in ensuring model trustworthiness.

\bibliographystyle{plain}
\bibliography{main}

\begin{thebibliography}{10}

\bibitem{Bengio1994}
Y.~Bengio, P.~Simard, and P.~Frasconi.
\newblock Learning long-term dependencies with gradient descent is difficult.
\newblock {\em IEEE Transactions on Neural Networks}, 5(2):157--166, 1994.

\bibitem{Courbariaux2016}
M.~Courbariaux, I.~Hubara, D.~Soudry, R.~El-Yaniv, and Y.~Bengio.
\newblock Binarized neural networks: Training deep neural networks with weights
  and activations constrained to +1 or -1.
\newblock {\em arXiv preprint arXiv:1602.02830}.

\bibitem{dosovitskiy2020image}
A.~Dosovitskiy et~al.
\newblock An image is worth 16x16 words: Transformers for image recognition at
  scale.
\newblock {\em arXiv preprint arXiv:2010.11929}, 2020.

\bibitem{Gupta2015}
S.~Gupta, A.~Agrawal, K.~Gopalakrishnan, and P.~Narayanan.
\newblock Deep learning with limited numerical precision.
\newblock In {\em Proceedings of the 32nd International Conference on Machine
  Learning (ICML-15)}, pages 1737--1746, 2015.

\bibitem{Han2015}
S.~Han, H.~Mao, and W.~J. Dally.
\newblock Deep compression: Compressing deep neural networks with pruning,
  trained quantization and huffman coding.
\newblock {\em arXiv preprint arXiv:1510.00149}.

\bibitem{He2015}
K.~He et~al.
\newblock Delving deep into rectifiers: Surpassing human-level performance on
  imagenet classification.
\newblock In {\em Proceedings of the IEEE international conference on computer
  vision}, pages 1026--1034, 2015.

\bibitem{he2016deep}
Kaiming He, Xiangyu Zhang, Shaoqing Ren, and Jian Sun.
\newblock Deep residual learning for image recognition.
\newblock In {\em Proceedings of the IEEE conference on computer vision and
  pattern recognition}, pages 770--778, 2016.

\bibitem{ieee754}
Institute of Electrical and Electronics Engineers.
\newblock {\em IEEE Standard for Floating-Point Arithmetic}, 2008.
\newblock Accessed: 2023-04-30.

\bibitem{batch}
S.~Ioffe and C.~Szegedy.
\newblock Batch normalization: Accelerating deep network training by reducing
  internal covariate shift.
\newblock In {\em Proceedings of the 32nd International Conference on Machine
  Learning}, pages 448--456, 2015.

\bibitem{kingma2014adam}
D.~P. Kingma and J.~Ba.
\newblock Adam: A method for stochastic optimization.
\newblock {\em arXiv preprint arXiv:1412.6980}, 2014.

\bibitem{maxpool}
A.~Krizhevsky, I.~Sutskever, and G.~E. Hinton.
\newblock Imagenet classification with deep convolutional neural networks.
\newblock In {\em Proceedings of the 25th International Conference on Neural
  Information Processing Systems - Volume 1}, pages 1097--1105, 2012.

\bibitem{krizhevsky2009learning}
Alex Krizhevsky.
\newblock Learning multiple layers of features from tiny images.
\newblock Technical report, Citeseer, 2009.

\bibitem{lecun1998gradient}
Yann LeCun, L{\'e}on Bottou, Yoshua Bengio, and Patrick Haffner.
\newblock Gradient-based learning applied to document recognition.
\newblock {\em Proceedings of the IEEE}, 86(11):2278--2324, 1998.

\bibitem{Micikevicius2017}
P.~Micikevicius et~al.
\newblock Mixed precision training.
\newblock {\em arXiv preprint arXiv:1710.03740}.

\bibitem{ruder2017overview}
S.~Ruder.
\newblock An overview of gradient descent optimization algorithms.
\newblock {\em arXiv preprint arXiv:1609.04747}, 2017.

\bibitem{Rumelhart1986}
D.~E. Rumelhart, G.~E. Hinton, and R.~J. Williams.
\newblock Learning representations by back-propagating errors.
\newblock {\em Nature}, 323(6088):533--536, 1986.

\bibitem{tieleman2012lecture}
T.~Tieleman and G.~Hinton.
\newblock Lecture 6.5-rmsprop: Divide the gradient by a running average of its
  recent magnitude.
\newblock COURSERA: Neural Networks for Machine Learning, 2012.

\bibitem{yun2023defense}
J.~Yun et~al.
\newblock In defense of pure 16-bit floating-point neural networks.
\newblock {\em arXiv preprint arXiv:2305.10947}, 2023.

\end{thebibliography}

\end{document}